\documentclass[letterpaper, 10 pt, conference]{ieeeconf}  

\IEEEoverridecommandlockouts                              

\usepackage{graphics} 
\usepackage{epsfig} 
\usepackage{mathptmx} 
\usepackage{times} 
\usepackage{amsmath} 
\usepackage{amssymb}  
\usepackage{multirow}
\usepackage{multicol}

\usepackage{bbding}

\usepackage[utf8]{inputenc} 
\usepackage[T1]{fontenc}    
\usepackage{hyperref}       
\usepackage{url}            
\usepackage{booktabs}       
\usepackage{amsfonts}       
\usepackage{nicefrac}       
\usepackage{microtype}      
\usepackage{xcolor}         
\usepackage{cleveref}
\usepackage{graphicx}
\usepackage{float}
\usepackage{multirow}
\usepackage{diagbox}

\usepackage{algorithm}
\usepackage{algpseudocode}
\usepackage{arydshln}

\usepackage{multirow}
\usepackage{listings}
\definecolor{dkgreen}{rgb}{0,0.6,0}
\definecolor{gray}{rgb}{0.5,0.5,0.5}
\definecolor{mauve}{rgb}{0.58,0,0.82}
\title{\LARGE \bf
Articulated Object Manipulation with Coarse-to-fine Affordance for Mitigating the Effect of Point Cloud Noise
}

\author{
Suhan Ling*, Yian Wang*, Shiguang Wu, Yuzheng Zhuang, Tianyi Xu, Yu Li, Chang Liu, Hao Dong
\thanks{Suhan Ling, Yian Wang, Tianyi Xu, Yu Li, Chang Liu, Hao Dong are with Hyperplane Lab, School of CS, Peking University and National Key Laboratory for Multimedia Information Processing. Shiguang Wu and Yuzheng Zhuang are with Huawei.} 
\thanks{* indicates equal contribution}
\thanks{
Corresponding to hao.dong@pku.edu.cn}%
}

\begin{document}
\maketitle

\begin{abstract} 
    3D articulated objects are inherently challenging for manipulation due to the varied geometries and intricate functionalities associated with articulated objects.
    Point-level affordance, which predicts the per-point actionable score and thus proposes the best point to interact with, has demonstrated excellent performance and generalization capabilities in articulated object manipulation. However, a significant challenge remains: while previous works use perfect point cloud generated in simulation, the models cannot directly apply to the noisy point cloud in the real-world. To tackle this challenge, we leverage the property of real-world scanned point cloud that, the point cloud becomes less noisy when the camera is closer to the object. Therefore, we propose a novel coarse-to-fine affordance learning pipeline to mitigate the effect of point cloud noise in two stages. In the first stage, we learn the affordance on the noisy far point cloud which includes the whole object to propose the approximated place to manipulate. Then, we move the camera in front of the approximated place, scan a less noisy point cloud containing precise local geometries for manipulation, and learn affordance on such point cloud to propose fine-grained final actions. The proposed method is thoroughly evaluated both using large-scale simulated noisy point clouds mimicking real-world scans, and in the real world scenarios, with superiority over existing methods, demonstrating the effectiveness in tackling the noisy real-world point cloud problem.

\end{abstract}

\section{INTRODUCTION}

    It is essential for next-generation robots to effectively assist humans and achieve precise interactions with common 3D articulated objects like cabinets and drawers. Unlike humans, robots lack innate abilities to understand part semantics, which makes it challenging to interact with highly articulated objects.
%

    Recent research has ventured into the realm of fine-grained manipulation affordance analysis based on 3D geometric inputs~\cite{deng20213d, varadarajan2012afnet, borja2022affordance, ju2024robo}. Notably, point-level affordance, which focuses on the geometric information of object local parts, and represents the per-point actionable information for manipulating diverse kinds of objects, has demonstrated its excellent performance and generalization abilities in various downstream tasks, including articulated object manipulation~\cite{Mo_2021_ICCV, wu2022vatmart, wang2022adaafford}, bimanual collaboration~\cite{zhao2022dualafford}, environment-aware manipulation~\cite{wu2023learningenv} and even deformable object manipulation~\cite{wu2023learning}.
    
    \begin{figure}[h]
    \begin{center}
        \includegraphics[width=0.95\linewidth]{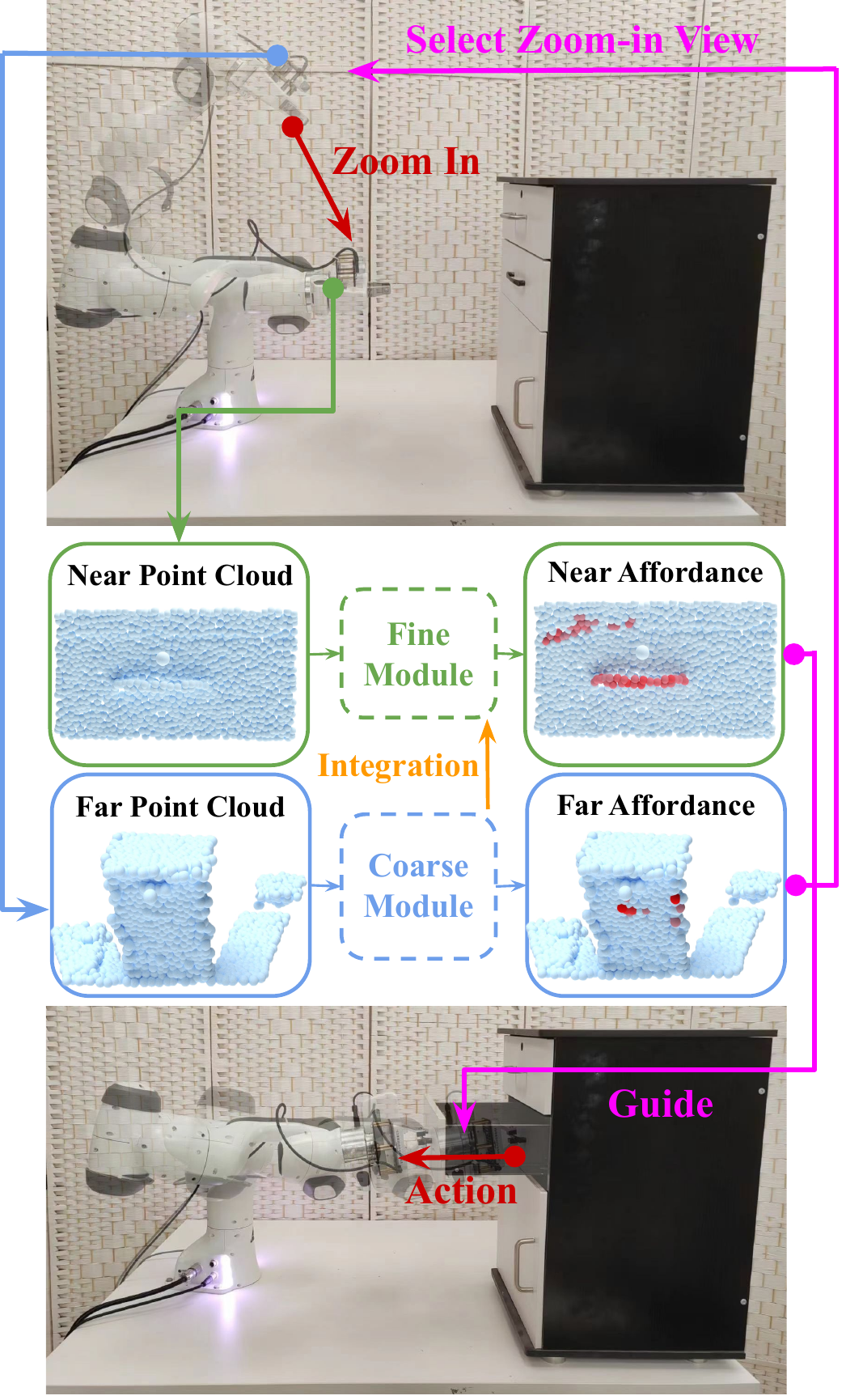}
    \end{center}
    \vspace{-5mm}
    \caption{Our proposed coarse-to-fine affordance learning framework articulated object manipulation with real-world noisy observations.}
    \vspace{-5mm}
    \label{fig:teasor}
    \end{figure}

However, when it comes to the practical application of manipulation in the real world, the above affordance learning approaches encounter the challenge, known as the sim-to-real gap. To be specific, policies may perform admirably in simulators that generate large-scale perfect point cloud,  but when directly deploying on the real-world scanned noisy point cloud, the policies trained in simulation will easily fail. The performance degradation from simulation to real world mainly comes from two reasons. First, the model trained only on perfect point cloud in simulation does not have the robustness to the out-of-distribution noisy point cloud in the real world. Second, the important geometries for manipulation, such as handles and edges may face heavy distortions and even disappear, cannot be reflected from the noisy point cloud, and thus the noisy point cloud cannot provide valid information for fine-grained manipulation.

To mitigate the effect of point-cloud noise for manipulation, this paper leverages the property that, the extent of noise will increase with the distance increase between the camera and the target object, and introduces a coarse-to-fine affordance learning framework. As is shown in Fig.~\ref{fig:teasor}, in the coarse stage, a coarse noisy point cloud in the far view is taken and the affordance prediction guides the approximated area to manipulate. Even though the significant local geometries such as handles are distorted or even disappear, the coarse affordance can estimate the rough positions to manipulate when taking the point cloud covering the whole shape as input. In the next fine stage, we move the camera mounted on the gripper to the front of the estimated manipulation area guided by the coarse affordance, and take a point cloud of the local area. The fine point cloud faces much slighter noise problem, and preserves the geometries of the target objects for manipulation. Besides, only taking the fine point cloud to propose actions possesses a limitation: it disregards global geometric context.  For example, when opening a door, the fine point cloud indicates where to manipulate, but does not contain information about the action direction, as the door axis will not be included in the point cloud in the near view. To address this problem, we introduce an integration in learning the affordance and proposing actions using a coarse-to-fine framework, by integrating the features of the coarse point cloud to the fine point cloud.

    
    In summary, our main contributions are the following.
    \begin{itemize}  
        \item 
        We propose a novel affordance framework that utilizes an eye-on-hand camera to obtain closer views tailored to the requirements of manipulation tasks. This approach effectively addresses the challenges posed by the noise present in the point cloud data.
        \item 
        We adapt PointNet++\cite{qi2017pointnetplusplus} to concurrently encode point cloud data captured by both the far and near view of the eye-on-hand camera. 
        \item 
        Experiments on noisy point clouds demonstrate our method can retain detailed geometry information from the closer point cloud and preserve global shape information from the farther.
    \end{itemize}

\section{RELATED WORK}
\label{sec:related_work}

    \subsection{3D Articulated Object Manipulation}
    Diverging from conventional 3D objects, articulated objects present intricacies in their kinematic properties attributed to varying joint types and joint limits. Consequently, a multitude of scholarly endeavors has arisen to delve into comprehending the structural dynamics of 3D articulated objects \cite{ nunes2019online, li2019articulated-pose, xu2022tandem, du2023learning} and how to manipulate them \cite{Mo_2021_ICCV, wu2022vatmart, wang2022adaafford, xu2022umpnet}. Comparing with grasping problems \cite{song2020grasping, akinola2021dynamic, fang2020graspnet, kokic2020learning, lenz2015deep, redmon2015real, qin2020s4g,bousmalis2018using}, manipulating 3D articulated object requires not only the detailed grasping pose but also how to execute interactions after grasping the correct point (\emph{e.g.}, when pulling open a door, the interaction trajectory should be an arc). Recent studies~\cite{wu2022vatmart, xu2022umpnet, eisner2022flowbot3d, schiavi2023learning, luo2023leverage} have explored the execution of sequential interactions to accomplish tasks, leveraging either known joint information or alternative approaches. While these investigations have demonstrated remarkable achievements in manipulating 3D articulated objects within simulated environments, their direct applicability to real-world scenarios remains challenging due to the significant sim-to-real gap.

    \subsection{Visual Affordance for 3D Shapes}
    Affordance, initially proposed by Gibson~\cite{gibson1977theory}, serves as a representation that conveys the interactive properties inherent in a scene~\cite{interaction-hotspots, interaction-exploration} or object~\cite{mandikal2020graff, corona2020ganhand} and has been successfully utilized in various studies~\cite{deng20213d, varadarajan2012afnet, varadarajan2012afrob, borja2022affordance}. In the context of manipulation tasks, the incorporation of visual affordance as an intermediate outcome can enhance the interpretability of the manipulation policies~\cite{wang2022adaafford, geng2022end, ning2023where2explore, wu2023learningenv}. Notably, the visual affordance heatmap provides a readily accessible means of visualizing the resulting affordance.

\section{Problem Formulation}
\label{sec:problem}

	Through observing the object from a relatively far view, our framework obtains a 3D partial point cloud $O^{far}\in\mathbb{R}^{N\times3}$ as input. Then our framework predicts an affordance score $a_{p}\in[0,1]$ for each single point $p\in\mathbb{R}^{3}$ in $O^{far}$, indicating the value of taking a closer view near this point, and choose the point with the highest score $p_{far} = argmax_{p} a_{p}$ to take a closer look at. After moving the camera closer to $p_{far}$, we scan another partial point cloud $O^{near}\in\mathbb{R}^{N\times3}$ on this view.

    Taking those two frames of point cloud as input, for each point $p\in O^{near}$ our framework proposes a set of action proposals $ACT_p=\{act_1, act_2, ...\}$ (\emph{i.e.}, different action orientations on the point $p$), namely $act_i=(p, R_i)$ where $p$ is the 3D point position and $R_i$ is the 6D gripper orientation. Here we generally follow the action definition of Where2act~\cite{Mo_2021_ICCV} by defining a task. 

    Finally, taking as input the two frames of point cloud $O^{far}$ and $O^{near}$ and the proposed action sets, our framework predicts the possibility of each action to successfully move the 3D articulation object and select the action with the highest score to execute. Specifically, we name this possibility as $c_{act}\in[0,1]$.


\section{Method}
\label{sec:method}

	\begin{figure*}[ht]
    \begin{center} 
        \includegraphics[width=\linewidth]{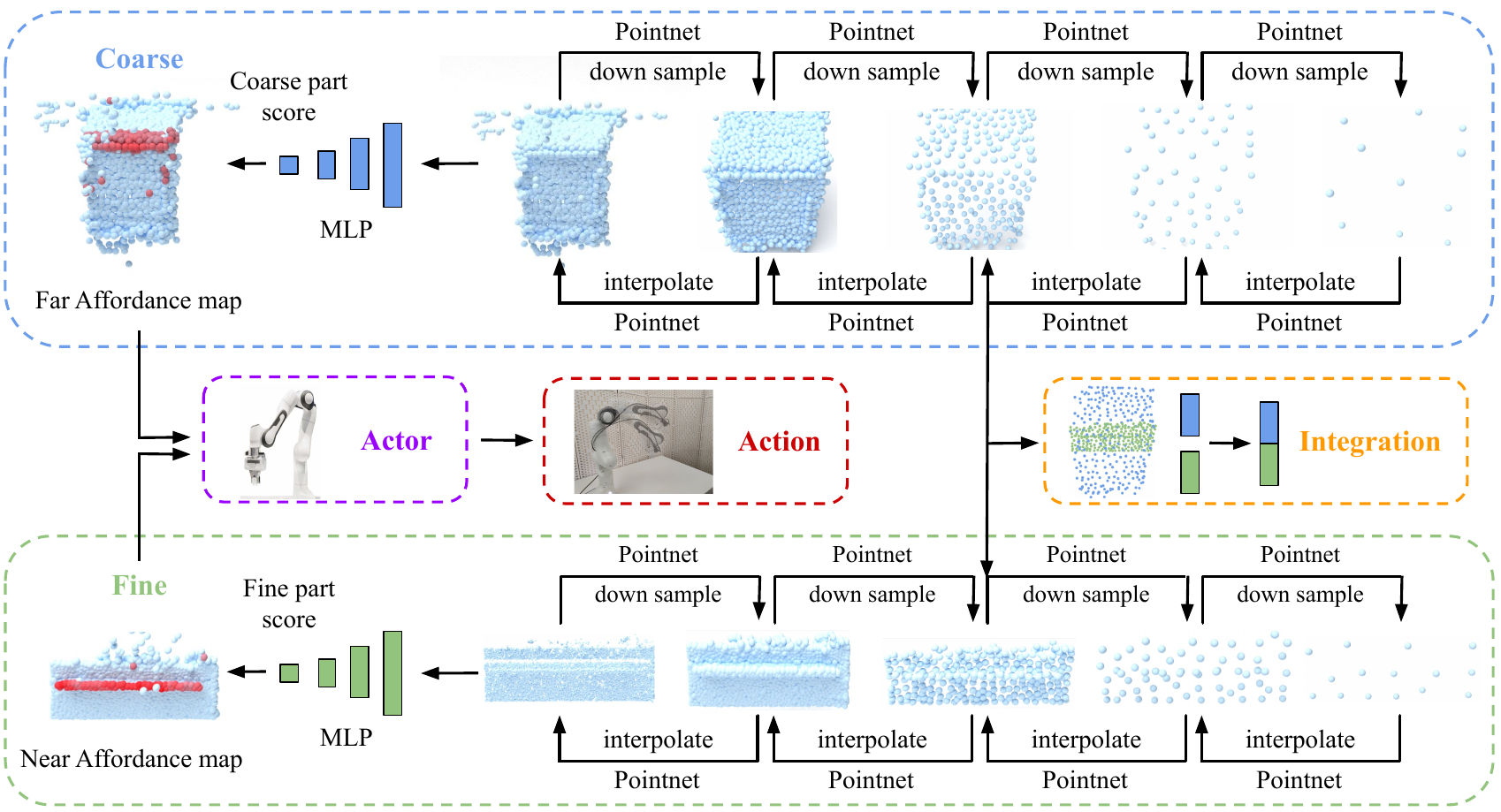}
    \end{center}
    \caption{Framework overview. Given the noisy point cloud in the far view as observation, our framework extracts per-point features using PointNet++ containing multi-scale pointnet, and then predicts a per-point coarse affordance map. We move the camera to the front of the point with the highest coarse affordance score, and take a less noisy point cloud. The framework uses another PointNet++ to extract per-point features of the fine point cloud, with the integration of the features of the far point cloud. The predicted affordance proposes the fine-grained actions.
    }
    \label{fig:method}
    \end{figure*}

    As shown in Fig~\ref{fig:method}, in general, our framework consists of three functional modules: 1) the \textit{Coarse} module takes only the far observation $O^{far}$ as input and predicts a coarse affordance score $a_{p}$ for each point $p\in O^{far}$. 2) the \textit{Fine} module takes the near observation $O^{near}$ and the action set $ACT_p$ for each $p\in O^{near}$ as input and predicts the feasibility $c_{act}$ of each action. Lastly, 3) the \textit{Actor} takes both $O^{far}$ and $O^{near}$ as input and outputs an action set for each point in $O^{near}$ for manipulating the object.

    To concurrently utilize both views of point cloud data, we make a special design in PointNet++ network~\cite{qi2017pointnetplusplus} to make an integration among our \textit{Coarse} module and \textit{Fine} module so the \textit{Fine} module can perceive global (\emph{i.e.}, far view) and local (\emph{i.e.}, near view) information as a whole. Below, we expound upon the intricate facets of each element within our pipeline.

    \subsection{Far-view Coarse Affordance}

    Despite the fact that the far view tends to receive much noise, it also captures crucial global information for selecting a nearer viewing position for closer inspection of objects. Therefore, a stationary far camera captures an object's comprehensive view and geometric overview. This information aids in assessing point affordance (\emph{i.e.}, $a_{p}$) and selecting the focal point for zooming in, our point of interest.
    
     Given the far view point cloud of the object, we utilize PointNet++ to regress $a_{p}$ for each point in $O^{far}$. The challenge lies in determining the ground-truth $a_{p}^{gt}$ for each point, since directly defining the ground-truth score as the success rate after taking this view is inadequate, as a result of the considerable object visibility in the closer view. 
    
    Assuming flawless \textit{Actor} and \textit{Fine} modules, interaction success hinges on the presence of an actionable point in the closer view, rendering local geometry inconsequential. Consequently, through trial-and-errors in simulated environment we set $a_{p}^{gt}=1$ if an interaction succeeds in the middle of the closer view, and $a_{p}^{gt}=0$ otherwise. This approach enhances the influence of local geometry on the predicted scores in the far view.

    \subsection{Near-view Fine Affordance}

    Although the far view generates a considerable amount of noise and is inadequate for accurate manipulation, it provides us a point of interest (\emph{i.e.}, the zoom-in point). In order to make the best of the global information provided by the far view while minimizing camera noise, the eye-on-hand camera is subsequently positioned at the point of interest, providing a near view of the target object. The near view point cloud data (\emph{i.e.}, $O^{near}$) acquired by such a camera is part of the input of our \textit{Fine} module, which functions as a critic network, assessing the feasibility of proposed actions of the \textit{Actor} module. Given $O^{near}$ as input, it employs PointNet++ to encode geometric attributes for each point $p\in O^{near}$, resulting in the point's geometry feature $f_p\in\mathbb{R}^{128}$. Subsequently, an MLP network integrates the geometry feature $f_p$ with action $act_p$, predicting success probability $c_{act}$.

    However, this \textit{Fine} module possesses a limitation: it disregards global geometric context. Consequently, it lacks 1) comprehensive awareness of boundary geometry in the closer view, and 2) consideration of vital joint information essential for manipulating articulated objects. To address this, we introduce an integration between the \textit{Coarse} and \textit{Fine} modules, enabling the transfer of global geometric insights from $O^{far}$ to the \textit{Fine} section.

    \subsection{Coarse-and-fine Integration}

    In order to fully utilize both far and near view information, intuitively these two views should be aligned and unified under the same world frame for further manipulation. Therefore, to establish a connection between the \textit{Coarse} and \textit{Fine} modules, we implement a unique design leveraging PointNet++. This design enables simultaneous encoding of $O^{near}$ and $O^{far}$, facilitating the transfer of global information from $O^{far}$ to $O^{near}$ while preserving local, detailed geometric information.

    In practical terms, as depicted in Fig.~\ref{fig:method}, we employ two separate PointNet++ encoders for $O^{far}$ and $O^{near}$. The latter layers of PointNet++ contain global insights, whereas the former layers retain intricate local details. To bridge the gap, we employ PointNet++'s Feature Propagation (FP) modules. Specifically, we interpolate the later layer's output onto the former layer, concatenate this interpolated feature with the feature from the former layer, and then process the concatenated feature through a unit point net to derive the final feature representation.

    \subsection{Actor Module}

    To manipulate the object, our \textit{Actor} module accounts for proposing feasible actions (\emph{i.e.}, the initial pulling or pushing direction), given the manipulation point.
    Specifically, we utilize a conditional Variational Autoencoder (cVAE) to propose suitable actions $act_p$ based on the geometric characteristics of a given point $p\in O^{near}$.

    \subsection{Implementation Details}

    Our data collection aims to capture effective manipulation dynamics, vital for training and our framework. We focus on creating a comprehensive dataset that robustly represents object-action interactions for point cloud affordance learning.

    Our systematic approach centers on interaction instances denoted as $D={O^{far}, p_{far}, O^{near}, p, act_p, gt}$. Here, $p_{far}$ signifies the examined point from $O^{far}$, while $p$ is the manipulation point within $O^{near}$. $act_p$ represents the action, and $gt\in{0, 1}$ indicates success based on exceeding a threshold $\tau$.

    Data collection efficiency is enhanced with Where2Act~\cite{mo2021where2act} pulling networks. Randomly sampling actionable points from point clouds in pulling tasks is inefficient due to high failure rates. By using the pre-trained Where2Act network, we streamline $p_{far}$, $p$, and $act_p$ selection, drastically reducing collection time. This allows for a larger dataset, significantly boosting network model performance.

    During training, we first jointly train the \textit{Coarse} and \textit{Fine} modules using positive and negative data. Subsequently, we train the \textit{Actor} module with positive data and the PointNet++ network from the \textit{Coarse} and \textit{Fine} modules.


    \textbf{Loss for the \textit{Coarse} module. }
    As outlined in section 4.1, when $p$ is centrally located in the near view, we use $a_{p}^{gt}=gt$ for direct supervision of the \textit{Coarse} module training. The near view is a closer look at $p_{far}$, and defining its "middle" pertains to the surrounding area of $p_{far}$. Therefore, we apply $a_{p}^{gt}=gt$ supervision when the distance between $p$ and $p_{far}$ is below a set threshold.

    A standard binary cross entropy loss is utilized here to measure the loss between the predicted score $a_p$ and its ground truth $a_{p}^{gt}$. The loss is defined as:
    \begin{equation}
    \mathcal{L}_{\textit{Coarse}}=-(a_{p}^{gt}\log(a_p) + (1-a_{p}^{gt})\log(1-a_p))\nonumber
    \end{equation}

    \textbf{Loss for the \textit{Fine} module. }
    Similar to the loss for the \textit{Coarse} module, given a piece of data $D$, and the prediction of the \textit{Fine} module $c_{act}$, we directly define its ground truth $c_{act}^{gt}=gt$. Then use a standard binary cross entropy loss to train the network,
    \begin{equation}
    \mathcal{L}_{\textit{Fine}}=-(c_{act}^{gt}\log(c_{act}) + (1-c_{act}^{gt})\log(1-c_{act}))\nonumber
    \end{equation}

    \textbf{Loss for the \textit{Actor} module. }
    Following Where2act~\cite{Mo_2021_ICCV}, which uses an action scoring module $D_s$ and $s_{R|p}=D_s(f_p, R)>0.5$ indicates a positive action proposal $R$, for a batch of $B$ interactions with their encoded point features $\{(f_{p_{i}},act_{p_{i}}, r_i)\}$, $r_i=1/0$ denotes ground-truth interaction outcome, we let: 
    \begin{equation}
    \mathcal{L}_{\textit{Actor}}=-\frac{1}{B} \sum_{i} r_i \log (D_s(f_{p_i}, R_i))+(1-r_i)\log (1-D_i(f_{p_i}, R_i))\nonumber
    \end{equation}


\section{Experiments}
\label{sec:exps}

    \begin{figure*}[ht]
    \begin{center}
        \includegraphics[width=\linewidth]{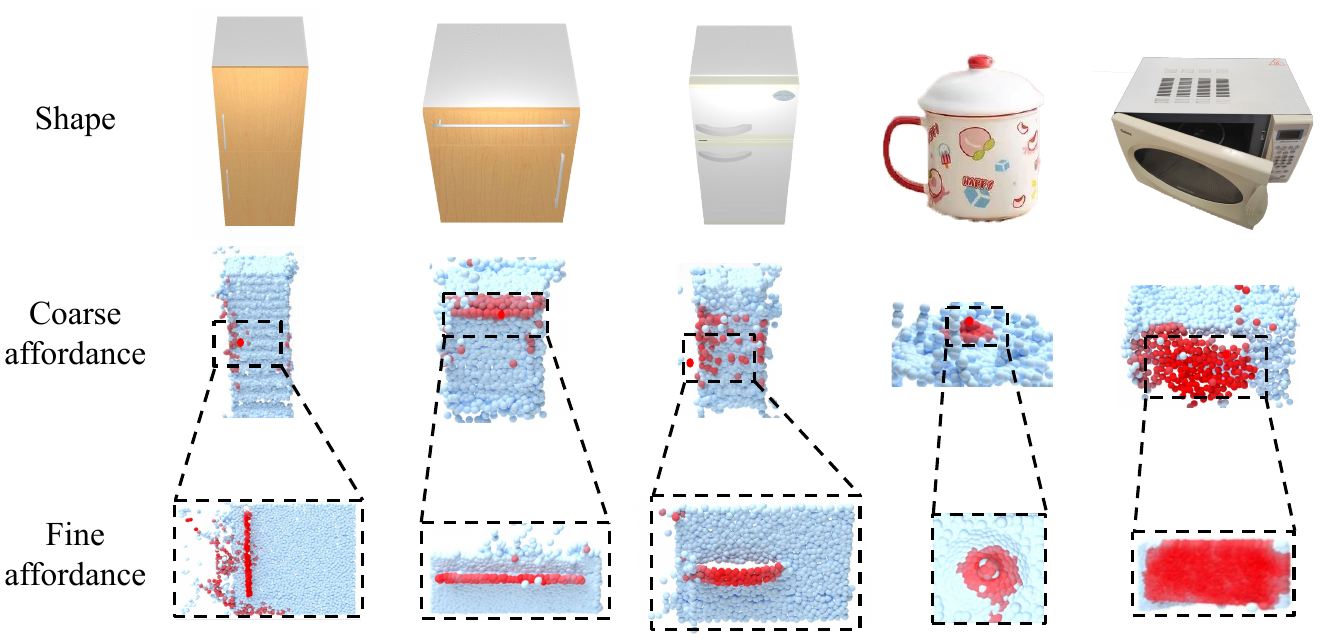}
    \end{center}
    \caption{Visualization of far view coarse affordance and near view fine affordance on noisy point clouds. The task for the microwave is ``push close'', while the task for others is ``pull open''. The first three shapes come from simulation, while the last two come from real-world scans.}
    \label{fig:visual}
    \end{figure*}

    This section covers the experimental setup, quantitative evaluations, and visual results, providing a comprehensive analysis of our proposed approach.

    \subsection{Experimental Setup}

    Using the SAPIEN physical simulator~\cite{Xiang_2020_SAPIEN} and the PartNet-Mobility dataset~\cite{Mo_2019_CVPR}, equipped with a ray-tracing depth camera \cite{Kuafu2022} that generates noisy point cloud with the same principle of real world sensors, our experiments rigorously evaluate our approach. Through visualizations and baseline comparisons, we validate our design's effectiveness both intuitively and quantitatively.

    \subsubsection{Environment}

    Building on Where2act, our novel approach simplifies robot arm complexities. Simulated object placement adheres to a zero-centered configuration, enhancing experimental control. We utilize a flying gripper or suctor based on categories for interactions, bypassing reachability challenges and focusing on object-centric policy learning. 

    Unlike previous methods using perfect point cloud, we utilize a ray-tracing depth camera \cite{Kuafu2022} for authentic point cloud generation in training and testing, which mitigates the sim-to-real disparities caused by unrealistic point clouds. The camera's role is two-fold: one is recording the distant view ($O^{far}$) from a predetermined position, and the other is capturing the nearby view ($O^{near}$) parallel to the ground.

    \subsubsection{Tasks}
    Inspired by VAT-Mart and built on Where2act's task and action design, we employ `PULL OPEN' and `PUSH CLOSE' as tasks. Moreover, we refine the pulling action for `PULL OPEN' in Where2act. Specifically, we separate the grasp pose from the subsequent movement, introducing a horizontal backward motion in the world frame for post-grasp movement. This adjustment streamlines our \textit{Fine} component by simplifying movement direction considerations, focusing on accurate grasping action assessment. 


    \subsubsection{Dataset}
    Our dataset comprises 8 categories, divided into 5 for training (StorageFurniture, Drawer, Microwave, Refrigerator, Kettle) and 3 for novel evaluation (WashingMachine, Pot, Safe). Each category is further divided into training and testing object sets. For instance, consider the \textit{StorageFurniture} category: the 345 objects are divided into a training set of 270 training objects and 75 testing objects.

    To gather interaction data, our process involves several steps. First, we select an object from the training set, position it in the scene, and capture $O^{far}$ using the camera to take a picture at the far and predetermined location. Then, we randomly identify $p_{far}$ within $O^{far}$ and position the camera at the same horizontal location as $p_{far}$ but 0.6 units ahead for scanning $O^{near}$.
    The depth information vanishes if the camera is moved too close to the object, so we must keep a reasonable distance.
    After the manipulation point $p$ is designated, we execute action $act_p$ and evaluate the result.

    For selecting $p_{far}$, $p$, and $act_p$, we use a strategy to boost positive data generation and speed up collection. This includes occasionally using pre-trained Where2act networks for decision-making, increasing favorable outcomes.

    Our dataset maintains equilibrium by equally using positive and negative data, each comprising half of the collection. Unlike prior methods, we exclude the need for part masks during testing. This simplification removes the burden of masking in real-world experiments, streamlining the workflow.

    \begin{table*}[tb]
\vspace{8mm}
 \begin{center}
   \caption{
   Comparisons of our method with baselines and ablations under ``pull open'' task.  
   }
 \small
     \setlength{\tabcolsep}{2mm}{
  \begin{tabular}{c|cccccc|cccc}
   \hline
 \textbf{Pull Open}
 &\multicolumn{6}{c|}{\textbf {Unseen Objects from Train Categories}}&\multicolumn{4}{c}{\textbf {Test Categories}}\\
   \hline
        Method
       &AVG
        &\includegraphics[width=0.04\linewidth]{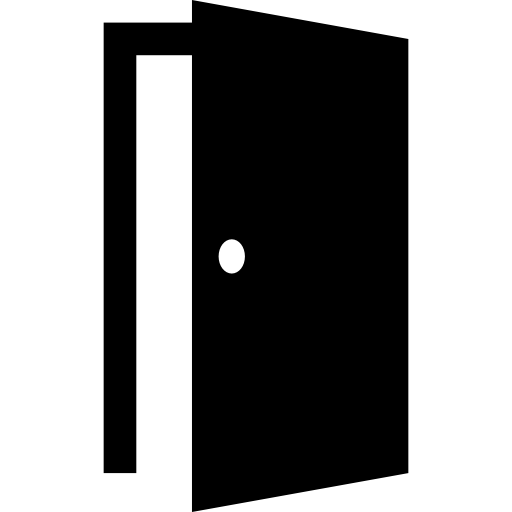}
        &\includegraphics[width=0.04\linewidth]{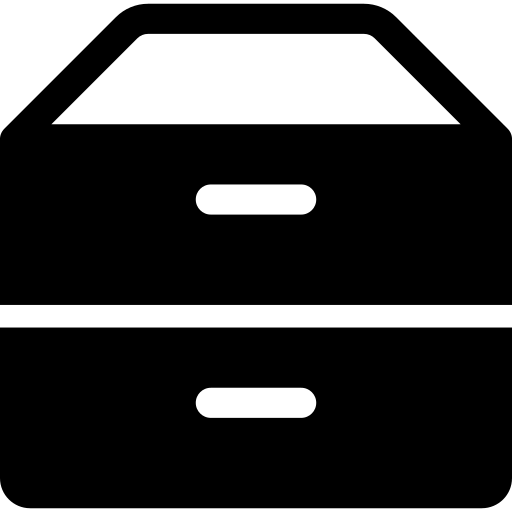}
        &\includegraphics[width=0.04\linewidth]{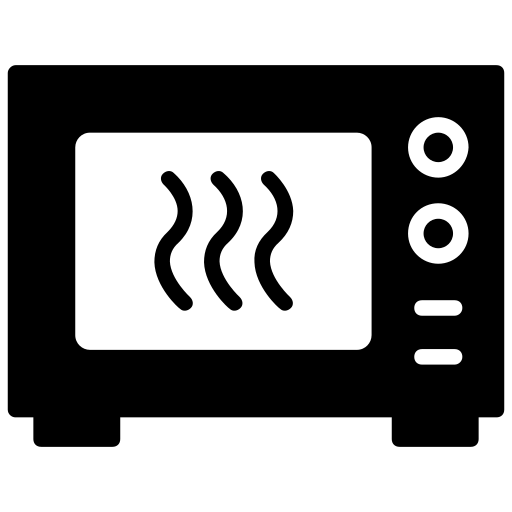}
        &\includegraphics[width=0.04\linewidth]{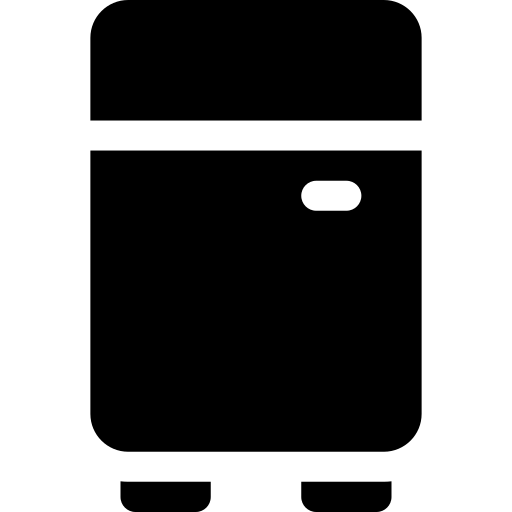}
        &\includegraphics[width=0.04\linewidth]{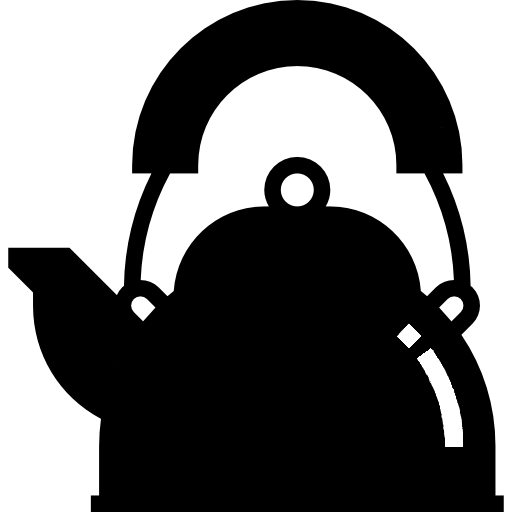}
        &AVG
        &\includegraphics[width=0.04\linewidth]{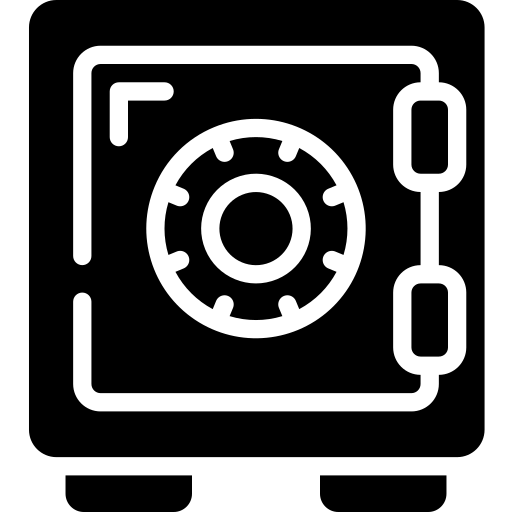}
        &\includegraphics[width=0.04\linewidth]{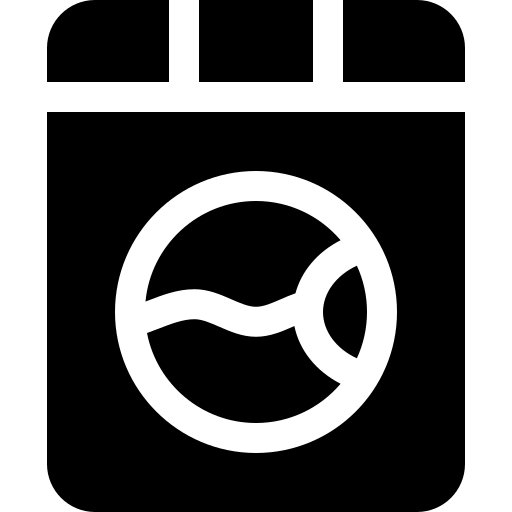}
        &\includegraphics[width=0.04\linewidth]{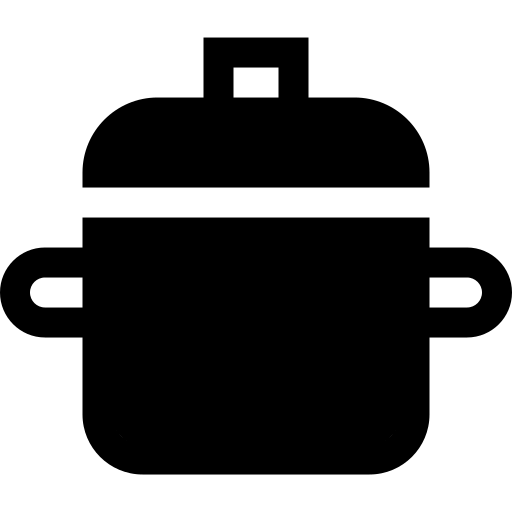} \\
  \hline\hline
  Where2Act & 0.21 & 0.11 & 0.16 & 0.33 & 0.27 & 0.18 & 0.21 & 0.25 & 0.22 & 0.15 \\
  FlowBot3D & 0.29 & 0.18 & 0.22 & 0.42 & 0.39 & 0.22 & 0.27 & 0.31 & 0.26 & 0.24 \\                              
  UMPNet & 0.35 & 0.25 & 0.29 & 0.41 & 0.49 & 0.29 & 0.32 & 0.36 & 0.31 & 0.28 \\
  VAT-Mart & 0.38 & 0.28 & 0.23 & 0.56 & 0.52 & 0.29 & 0.35 & 0.46 & 0.32 & 0.27 \\
  \hline
  Ours Random-Coarse & 0.33 & 0.35 & 0.25 & 0.38 & 0.48 & 0.21 & 0.24 & 0.31 & 0.22 & 0.20 \\
  Ours Random-Fine & 0.35 & 0.21 & 0.26 & 0.49 & 0.54 & 0.24 & 0.27 & 0.41 & 0.21 & 0.18 \\
  Ours Separate & 0.50 & 0.45 & 0.49 & 0.59 & 0.53 & 0.43 & 0.39 & 0.48 & 0.30 & 0.39 \\
  Ours Final & \textbf{0.61} & \textbf{0.58} & \textbf{0.63} & \textbf{0.68} & \textbf{0.64} & \textbf{0.51} & \textbf{0.50} & \textbf{0.56} & \textbf{0.48} & \textbf{0.45} \\
  \hline
  
  \end{tabular}}

  \label{tab:pull}
 \end{center}
\end{table*}

    \begin{table*}[tb]
 \begin{center}

   \caption{
   Comparisons of our method with baselines and ablations under ``push close'' task.  
   }
   
 \small
     \setlength{\tabcolsep}{2mm}{
  \begin{tabular}{c|cccccc|cccc}
   \hline
   \textbf{Push Close}
   &\multicolumn{6}{c|}{\textbf {Unseen Objects from Train Categories}}&\multicolumn{4}{c}{\textbf {Test Categories}}\\
   \hline
 Method
        &AVG
        &\includegraphics[width=0.04\linewidth]{tabs/images/door.png}
        &\includegraphics[width=0.04\linewidth]{tabs/images/drawers.png}
        &\includegraphics[width=0.04\linewidth]{tabs/images/microwave-oven.png}
        &\includegraphics[width=0.04\linewidth]{tabs/images/fridge.png}
        &\includegraphics[width=0.04\linewidth]{tabs/images/kettle.png}
        &AVG
        &\includegraphics[width=0.04\linewidth]{tabs/images/safe.png}
        &\includegraphics[width=0.04\linewidth]{tabs/images/washing-machine.png}
        &\includegraphics[width=0.04\linewidth]{tabs/images/pot.png} \\
  \hline\hline
  Where2Act & 0.80 & 0.81 & 0.75 & 0.85 & 0.81 & 0.77 & 0.66 & 0.66 & 0.52 & 0.80 \\
  FlowBot3D & 0.86 & 0.90 & 0.84 & 0.85 & 0.88 & 0.81 & 0.69 & 0.72 & 0.58 & 0.78 \\                              
  UMPNet & 0.84 & 0.83 & 0.86 & 0.88 & 0.82 & 0.83 & 0.75 & 0.78 & 0.63 & 0.84 \\
  VAT-Mart & 0.91 & 0.93 & 0.93 & 0.91 & 0.92 & 0.87 & 0.77 & 0.83 & 0.59 & 0.89 \\
  \hline
  Ours Random-Coarse & 0.43 & 0.50 & 0.48 & 0.39 & 0.47 & 0.33 & 0.40 & 0.40 & 0.27 & 0.53 \\
  Ours Random-Fine & 0.79 & 0.86 & 0.83 & 0.75 & 0.80 & 0.71 & 0.53 & 0.76 & 0.21 & 0.61 \\
  Ours Separate & 0.92 & 0.91 & 0.94 & 0.96 & 0.91 & 0.90 & 0.71 & 0.74 & 0.57 & 0.83 \\
  Ours Final & \textbf{0.96} & \textbf{0.95} & \textbf{0.98} & \textbf{0.98} & \textbf{0.95} & \textbf{0.93} & \textbf{0.85} & \textbf{0.88} & \textbf{0.71} & \textbf{0.96} \\
  \hline
  
  \end{tabular}}
  \vspace{-0.1cm}

  \label{tab:push}
 \end{center}
\end{table*}

    \vspace{-1mm}
    \subsection{Baselines, Ablations and Evaluation Metrics}
\vspace{-1mm}
    We conduct at least 200 interactions in random test shapes and calculate successful rates.


    For a fair comparison, we have constrained the scope of the baseline studies to the affordance method, thereby highlighting the superiority of our approach. Specifically, we compare against four baselines (Where2act~\cite{mo2021where2act}, FlowBot3D~\cite{eisner2022flowbot3d}, UMPNet~\cite{xu2022universal}, VAT-Mart~\cite{wu2022vatmart}). We retain their architectures, replacing training data with our realistic point cloud data and tasks, showcasing the strength of our approach.

    We also implement three ablations to demonstrate the significance of coarse affordance, fine affordance and coarse-to-fine integration.
    
    \begin{itemize}
    
        \item \textbf{Ours Random-Coarse}: We eliminate the \textit{Coarse} module and choose $p_{far}$ randomly.

        \item \textbf{Ours Random-Fine}: We eliminate the \textit{Fine} module and choose $p_{fine}$ randomly.

        \item \textbf{Ours Separate}: In the \textit{Fine} module, instead of making the integration using interpolation, we first encode $O^{far}$ into a vector $z$ as an extra input along with the point features $f_p$.
    
    \end{itemize}

    The first two are used to demonstrate the importance of the \textit{Coarse} module (\emph{i.e.}, the benefits of choosing a zoom-in point with the highest affordance score generated by the far view) and the \textit{Fine} module 
    while the last one is used to exhibit the effectiveness of our coarse-to-fine integration design.
    
    
    \subsection{Results and Analysis}
    Tab.~\ref{tab:pull} and ~\ref{tab:push} show the numerical comparisons with baselines and ablations,
    and Fig.~\ref{fig:visual} shows the visualization of coarse and fine affordance on noisy point cloud generated in simulation or scanned in the real world.

    In our comparative analysis, we emphasize the effectiveness and efficiency of our key components: coarse affordance, fine affordance, zoom-in policy, and the integration of \textit{Coarse} and \textit{Fine} modules. These elements collectively shape the success and performance of our framework.

    Shown in Fig~\ref{fig:visual}, \textit{Coarse} module captures overall object geometry and proposes reasonable zoom-in points. Meanwhile, \textit{Fine} module refines results using detailed geometry.

    In zoom-in policy evaluation, we compare with currently most strong baselines, while are in lack of zoom-in. Our model outperforms all baselines, as the baselines rely heavily on detailed geometries, while noisy far view point cloud could not reflect fine-grained geometries.

Results in Table~\ref{tab:pull} and ~\ref{tab:push} highlight the effectiveness of our coarse and fine affordance respectively. Comparing \emph{Ours-Final} to \emph{Ours-Random-Coarse} and \emph{Ours-Random-Fine} ablations, our approach, selecting points strategically based on coarse affordance, and then selecting points strategically based on fine affordance, outperforms respective random point selection. This reinforces that our method improves interaction efficiency by intelligently prioritizing points with strong coarse affordance and then fine affordance.
   
    Comparing \emph{Ours-Final} with the \emph{Ours-Separate} ablation provides compelling evidence for our integration design's necessity and effectiveness. Our approach unifies far and near affordance encoding, while the ablation treats them separately. The results convincingly reveal that the performance could be enhanced through improved information coherence and integration, showcasing our pipeline's robustness.

    \begin{table}[tb]
 \begin{center}
   \caption{Success Rate of Manipulation in the Real World.}
   
 \small
     \setlength{\tabcolsep}{2mm}{
  \begin{tabular}{c|cccc|cc}
   \hline
 \multirow{2}{*}{\textbf{}}&\multicolumn{4}{c|}{\textbf{pull open}}&\multicolumn{2}{c}{\textbf{push close}}\\
   \hline
        Method
        &\includegraphics[width=0.04\linewidth]{tabs/images/door.png}
        &\includegraphics[width=0.04\linewidth]{tabs/images/drawers.png}
        &\includegraphics[width=0.04\linewidth]{tabs/images/kettle.png}
        &\includegraphics[width=0.04\linewidth]{tabs/images/pot.png}
        &\includegraphics[width=0.04\linewidth]{tabs/images/drawers.png}&\includegraphics[width=0.04\linewidth]{tabs/images/pot.png} \\
  \hline\hline
  FlowBot3D &3/10  &2/10  &3/10  &2/10 & 7/10  &6/10  \\
  VAT-Mart &4/10 &4/10 &2/10 &3/10 &7/10 &8/10 \\
  Ours&\textbf{6/10}&\textbf{7/10}&\textbf{6/10}&\textbf{5/10}&\textbf{9/10}&\textbf{10/10} \\
  \hline
  
  \end{tabular}}

  \label{tab:real}
 \end{center}
 \vspace{-0.55cm}
\end{table}

    \vspace{-2mm}
    \subsection{Real-world Experiment}
\vspace{-1mm}
    To verify our method's real-world applicability and its ability, we conduct experiments using a 7-DOF Franka Emika Panda robot with a Realsense camera mounted on its hand. The full setup can be seen in Fig.~\ref{fig:teasor}.
    
    In real-world experiments, we sequentially capture far and near point cloud to generate the affordance map as the object-centric representation which provides us the manipulation point and interaction direction. To achieve long-term manipulation, we employ a heuristic method to sequentially output the next action and execute it with cartesian impedance control which transforms the end-effector action to the robot.

    We conduct experiments on different categories of articulated objects, and conduct 10 interactions on each object. Results in Fig.~\ref{fig:visual} and Tab.~\ref{tab:real} demonstrate our method can simply apply to real world without any other techniques.


\vspace{-3mm}
\section{Conclusion}
\vspace{-3mm}

    In conclusion, our study presents a novel coarse-to-fine pipeline that effectively utilizes both far and near point clouds, integrating them to mitigate the noisy point cloud problem. Through comprehensive evaluations, we have demonstrated that our method surpasses the baseline and ablations across all 8 categories considered. Furthermore, our approach, by leveraging an affordance method, exhibits a favorable generalizing ability on novel shapes and categories, indicating its potential for real-world applications. It is worth noting that our method requires only commercially available cameras, making it accessible for widespread adoption. 




\vspace{-2mm}
\section*{ACKNOWLEDGMENT}
\vspace{-2mm}

This project was supported by The National Youth Talent Support Program (8200800081) and National Natural Science Foundation of China (No. 62136001).


{
\bibliographystyle{IEEEtran}
\bibliography{IEEEabrv,references}

\begin{thebibliography}{10}
\providecommand{\url}[1]{#1}
\csname url@samestyle\endcsname
\providecommand{\newblock}{\relax}
\providecommand{\bibinfo}[2]{#2}
\providecommand{\BIBentrySTDinterwordspacing}{\spaceskip=0pt\relax}
\providecommand{\BIBentryALTinterwordstretchfactor}{4}
\providecommand{\BIBentryALTinterwordspacing}{\spaceskip=\fontdimen2\font plus
\BIBentryALTinterwordstretchfactor\fontdimen3\font minus
  \fontdimen4\font\relax}
\providecommand{\BIBforeignlanguage}[2]{{%
\expandafter\ifx\csname l@#1\endcsname\relax
\typeout{** WARNING: IEEEtran.bst: No hyphenation pattern has been}%
\typeout{** loaded for the language `#1'. Using the pattern for}%
\typeout{** the default language instead.}%
\else
\language=\csname l@#1\endcsname
\fi
#2}}
\providecommand{\BIBdecl}{\relax}
\BIBdecl

\bibitem{deng20213d}
S.~Deng, X.~Xu, C.~Wu, K.~Chen, and K.~Jia, ``3d affordancenet: A benchmark for
  visual object affordance understanding,'' in \emph{proceedings of the
  IEEE/CVF conference on computer vision and pattern recognition}, 2021, pp.
  1778--1787.

\bibitem{varadarajan2012afnet}
K.~M. Varadarajan and M.~Vincze, ``Afnet: The affordance network,'' in
  \emph{Asian conference on computer vision}.\hskip 1em plus 0.5em minus
  0.4em\relax Springer, 2012, pp. 512--523.

\bibitem{borja2022affordance}
J.~Borja-Diaz, O.~Mees, G.~Kalweit, L.~Hermann, J.~Boedecker, and W.~Burgard,
  ``Affordance learning from play for sample-efficient policy learning,'' in
  \emph{2022 International Conference on Robotics and Automation (ICRA)}.\hskip
  1em plus 0.5em minus 0.4em\relax IEEE, 2022, pp. 6372--6378.

\bibitem{ju2024robo}
Y.~Ju, K.~Hu, G.~Zhang, G.~Zhang, M.~Jiang, and H.~Xu, ``Robo-abc: Affordance
  generalization beyond categories via semantic correspondence for robot
  manipulation,'' \emph{arXiv preprint arXiv:2401.07487}, 2024.

\bibitem{Mo_2021_ICCV}
K.~Mo, L.~J. Guibas, M.~Mukadam, A.~Gupta, and S.~Tulsiani, ``Where2act: From
  pixels to actions for articulated 3d objects,'' in \emph{Proceedings of the
  IEEE/CVF International Conference on Computer Vision (ICCV)}, October 2021,
  pp. 6813--6823.

\bibitem{wu2022vatmart}
\BIBentryALTinterwordspacing
R.~Wu, Y.~Zhao, K.~Mo, Z.~Guo, Y.~Wang, T.~Wu, Q.~Fan, X.~Chen, L.~Guibas, and
  H.~Dong, ``{VAT}-mart: Learning visual action trajectory proposals for
  manipulating 3d {ART}iculated objects,'' in \emph{International Conference on
  Learning Representations}, 2022. [Online]. Available:
  \url{https://openreview.net/forum?id=iEx3PiooLy}
\BIBentrySTDinterwordspacing

\bibitem{wang2022adaafford}
Y.~Wang, R.~Wu, K.~Mo, J.~Ke, Q.~Fan, L.~Guibas, and H.~Dong, ``{AdaAfford}:
  Learning to adapt manipulation affordance for 3d articulated objects via
  few-shot interactions,'' \emph{European conference on computer vision}, 2022.

\bibitem{zhao2022dualafford}
Y.~Zhao, R.~Wu, Z.~Chen, Y.~Zhang, Q.~Fan, K.~Mo, and H.~Dong, ``Dualafford:
  Learning collaborative visual affordance for dual-gripper object
  manipulation,'' \emph{International Conference on Learning Representations
  (ICLR)}, 2023.

\bibitem{wu2023learningenv}
\BIBentryALTinterwordspacing
R.~Wu, K.~Cheng, Y.~Zhao, C.~Ning, G.~Zhan, and H.~Dong, ``Learning
  environment-aware affordance for 3d articulated object manipulation under
  occlusions,'' in \emph{Thirty-seventh Conference on Neural Information
  Processing Systems}, 2023. [Online]. Available:
  \url{https://openreview.net/forum?id=Re2NHYoZ5l}
\BIBentrySTDinterwordspacing

\bibitem{wu2023learning}
R.~Wu, C.~Ning, and H.~Dong, ``Learning foresightful dense visual affordance
  for deformable object manipulation,'' in \emph{IEEE International Conference
  on Computer Vision (ICCV)}, 2023.

\bibitem{qi2017pointnetplusplus}
C.~R. Qi, L.~Yi, H.~Su, and L.~J. Guibas, ``Pointnet++: Deep hierarchical
  feature learning on point sets in a metric space,'' \emph{arXiv preprint
  arXiv:1706.02413}, 2017.

\bibitem{nunes2019online}
U.~M. Nunes and Y.~Demiris, ``Online unsupervised learning of the 3d kinematic
  structure of arbitrary rigid bodies,'' in \emph{Proceedings of the IEEE/CVF
  International Conference on Computer Vision}, 2019, pp. 3809--3817.

\bibitem{li2019articulated-pose}
X.~Li, H.~Wang, L.~Yi, L.~Guibas, A.~L. Abbott, and S.~Song, ``Category-level
  articulated object pose estimation,'' \emph{arXiv preprint arXiv:1912.11913},
  2019.

\bibitem{xu2022tandem}
J.~Xu, S.~Song, and M.~Ciocarlie, ``Tandem: Learning joint exploration and
  decision making with tactile sensors,'' \emph{IEEE Robotics and Automation
  Letters}, 2022.

\bibitem{du2023learning}
Y.~Du, R.~Wu, Y.~Shen, and H.~Dong, ``Learning part motion of articulated
  objects using spatially continuous neural implicit representations,'' in
  \emph{British Machine Vision Conference (BMVC)}, November 2023.

\bibitem{xu2022umpnet}
Z.~Xu, H.~Zhanpeng, and S.~Song, ``Umpnet: Universal manipulation policy
  network for articulated objects,'' \emph{IEEE Robotics and Automation
  Letters}, 2022.

\bibitem{song2020grasping}
S.~Song, A.~Zeng, J.~Lee, and T.~Funkhouser, ``Grasping in the wild: Learning
  6dof closed-loop grasping from low-cost demonstrations,'' \emph{Robotics and
  Automation Letters}, 2020.

\bibitem{akinola2021dynamic}
I.~Akinola, J.~Xu, S.~Song, and P.~K. Allen, ``Dynamic grasping with
  reachability and motion awareness,'' in \emph{2021 IEEE/RSJ International
  Conference on Intelligent Robots and Systems (IROS)}.\hskip 1em plus 0.5em
  minus 0.4em\relax IEEE, 2021, pp. 9422--9429.

\bibitem{fang2020graspnet}
H.-S. Fang, C.~Wang, M.~Gou, and C.~Lu, ``Graspnet-1billion: A large-scale
  benchmark for general object grasping,'' in \emph{Proceedings of the IEEE/CVF
  Conference on Computer Vision and Pattern Recognition(CVPR)}, 2020, pp.
  11\,444--11\,453.

\bibitem{kokic2020learning}
M.~Kokic, D.~Kragic, and J.~Bohg, ``Learning task-oriented grasping from human
  activity datasets,'' \emph{IEEE Robotics and Automation Letters}, vol.~5,
  no.~2, pp. 3352--3359, 2020.

\bibitem{lenz2015deep}
I.~Lenz, H.~Lee, and A.~Saxena, ``Deep learning for detecting robotic grasps,''
  \emph{The International Journal of Robotics Research}, vol.~34, no. 4-5, pp.
  705--724, 2015.

\bibitem{redmon2015real}
J.~Redmon and A.~Angelova, ``Real-time grasp detection using convolutional
  neural networks,'' in \emph{2015 IEEE International Conference on Robotics
  and Automation (ICRA)}.\hskip 1em plus 0.5em minus 0.4em\relax IEEE, 2015,
  pp. 1316--1322.

\bibitem{qin2020s4g}
Y.~Qin, R.~Chen, H.~Zhu, M.~Song, J.~Xu, and H.~Su, ``S4g: Amodal single-view
  single-shot se (3) grasp detection in cluttered scenes,'' in \emph{Conference
  on robot learning}.\hskip 1em plus 0.5em minus 0.4em\relax PMLR, 2020, pp.
  53--65.

\bibitem{bousmalis2018using}
K.~Bousmalis, A.~Irpan, P.~Wohlhart, Y.~Bai, M.~Kelcey, M.~Kalakrishnan,
  L.~Downs, J.~Ibarz, P.~Pastor, K.~Konolige \emph{et~al.}, ``Using simulation
  and domain adaptation to improve efficiency of deep robotic grasping,'' in
  \emph{2018 IEEE international conference on robotics and automation
  (ICRA)}.\hskip 1em plus 0.5em minus 0.4em\relax IEEE, 2018, pp. 4243--4250.

\bibitem{eisner2022flowbot3d}
B.~Eisner, H.~Zhang, and D.~Held, ``Flowbot3d: Learning 3d articulation flow to
  manipulate articulated objects,'' \emph{arXiv preprint arXiv:2205.04382},
  2022.

\bibitem{schiavi2023learning}
G.~Schiavi, P.~Wulkop, G.~Rizzi, L.~Ott, R.~Siegwart, and J.~J. Chung,
  ``Learning agent-aware affordances for closed-loop interaction with
  articulated objects,'' in \emph{2023 IEEE International Conference on
  Robotics and Automation (ICRA)}.\hskip 1em plus 0.5em minus 0.4em\relax IEEE,
  2023, pp. 5916--5922.

\bibitem{luo2023leverage}
H.~Luo, W.~Zhai, J.~Zhang, Y.~Cao, and D.~Tao, ``Leverage interactive affinity
  for affordance learning,'' in \emph{Proceedings of the IEEE/CVF Conference on
  Computer Vision and Pattern Recognition}, 2023, pp. 6809--6819.

\bibitem{gibson1977theory}
J.~J. Gibson, ``The theory of affordances,'' \emph{Hilldale, USA}, vol.~1,
  no.~2, pp. 67--82, 1977.

\bibitem{interaction-hotspots}
T.~Nagarajan, C.~Feichtenhofer, and K.~Grauman, ``Grounded human-object
  interaction hotspots from video,'' in \emph{ICCV}, 2019.

\bibitem{interaction-exploration}
T.~Nagarajan and K.~Grauman, ``Learning affordance landscapes for interaction
  exploration in 3d environments,'' in \emph{NeurIPS}, 2020.

\bibitem{mandikal2020graff}
P.~Mandikal and K.~Grauman, ``Learning dexterous grasping with object-centric
  visual affordances,'' in \emph{IEEE International Conference on Robotics and
  Automation (ICRA)}, 2021.

\bibitem{corona2020ganhand}
E.~Corona, A.~Pumarola, G.~Alenya, F.~Moreno-Noguer, and G.~Rogez, ``Ganhand:
  Predicting human grasp affordances in multi-object scenes,'' in
  \emph{Proceedings of the IEEE/CVF Conference on Computer Vision and Pattern
  Recognition}, 2020, pp. 5031--5041.

\bibitem{varadarajan2012afrob}
K.~M. Varadarajan and M.~Vincze, ``Afrob: The affordance network ontology for
  robots,'' in \emph{2012 IEEE/RSJ international conference on intelligent
  robots and systems}.\hskip 1em plus 0.5em minus 0.4em\relax IEEE, 2012, pp.
  1343--1350.

\bibitem{geng2022end}
Y.~Geng, B.~An, H.~Geng, Y.~Chen, Y.~Yang, and H.~Dong, ``End-to-end affordance
  learning for robotic manipulation,'' \emph{arXiv preprint arXiv:2209.12941},
  2022.

\bibitem{ning2023where2explore}
C.~Ning, R.~Wu, H.~Lu, K.~Mo, and H.~Dong, ``Where2explore: Few-shot affordance
  learning for unseen novel categories of articulated objects,'' \emph{arXiv
  preprint arXiv:2309.07473}, 2023.

\bibitem{mo2021where2act}
K.~Mo, L.~J. Guibas, M.~Mukadam, A.~Gupta, and S.~Tulsiani, ``Where2act: From
  pixels to actions for articulated 3d objects,'' in \emph{Proceedings of the
  IEEE/CVF International Conference on Computer Vision}, 2021, pp. 6813--6823.

\bibitem{Xiang_2020_SAPIEN}
F.~Xiang, Y.~Qin, K.~Mo, Y.~Xia, H.~Zhu, F.~Liu, M.~Liu, H.~Jiang, Y.~Yuan,
  H.~Wang, L.~Yi, A.~X. Chang, L.~J. Guibas, and H.~Su, ``{SAPIEN}: A simulated
  part-based interactive environment,'' in \emph{The IEEE Conference on
  Computer Vision and Pattern Recognition (CVPR)}, June 2020.

\bibitem{Mo_2019_CVPR}
K.~Mo, S.~Zhu, A.~X. Chang, L.~Yi, S.~Tripathi, L.~J. Guibas, and H.~Su,
  ``{PartNet}: A large-scale benchmark for fine-grained and hierarchical
  part-level {3D} object understanding,'' in \emph{The IEEE Conference on
  Computer Vision and Pattern Recognition (CVPR)}, June 2019.

\bibitem{Kuafu2022}
A.~L. Xiaoshuai~Zhang, Fanbo~Xiang, ``Kuafu: A real-time ray tracing
  renderer,'' \url{https://github.com/jetd1/kuafu}, 2022, accessed: January 18,
  2023.

\bibitem{xu2022universal}
Z.~Xu, Z.~He, and S.~Song, ``Universal manipulation policy network for
  articulated objects,'' \emph{IEEE Robotics and Automation Letters}, vol.~7,
  no.~2, pp. 2447--2454, 2022.

\end{thebibliography}
}

\end{document}